\def\year{2019}\relax
\documentclass[letterpaper]{article} 
\usepackage{aaai19}  
\usepackage{times}  
\usepackage{helvet} 
\usepackage{courier}  
\usepackage[hyphens]{url}  
\usepackage[utf8]{inputenc}
\usepackage{graphicx} 
\usepackage{graphicx}  
\title{HireNet: a Hierarchical Attention Model for the Automatic Analysis of Asynchronous Video Job Interviews}

\author{
Léo Hemamou\textsuperscript{1,2,3},
Ghazi Felhi\textsuperscript{1},
Vincent Vandenbussche\textsuperscript{1},
Jean-Claude Martin\textsuperscript{2},
Chloé Clavel\textsuperscript{3}\\
\textsuperscript{1}{EASYRECRUE, 3 bis Rue de la Chaussée d'Antin, 75009 Paris, France}\\
\textsuperscript{2}{LIMSI, CNRS, Paris-Sud University, Paris-Saclay University / F-91405 Orsay, France}\\
\textsuperscript{3}{LTCI, Télécom ParisTech, Paris-Saclay University / F-75013 Paris, France}\\
\{l.hemamou,g.felhi,v.vandenbussche\}@easyrecrue.com,
Jean-Claude.Martin@limsi.fr,
chloe.clavel@telecom-paristech.fr
}

\begin{document}

\maketitle
\begin{abstract}

New technologies drastically change recruitment techniques. Some research projects aim at designing interactive systems that help candidates practice job interviews. Other studies aim at the automatic detection of social signals (\textit{e.g.} smile, turn of speech, etc...) in videos of job interviews. These studies are limited with respect to the number of interviews they process, but also by the fact that they only analyze simulated job interviews (\textit{e.g.} students pretending to apply for a fake position). Asynchronous video interviewing tools have become mature products on the human resources market, and thus, a popular step in the recruitment process. As part of a project to help recruiters, we collected a corpus of more than 7000 candidates having asynchronous video job interviews for real positions and recording videos of themselves answering a set of questions. We propose a new hierarchical attention model called HireNet that aims at predicting the hirability of the candidates as evaluated by recruiters. In HireNet, an interview is considered as a sequence of questions and answers containing salient socials signals. Two contextual sources of information are modeled in HireNet: the words contained in the question and in the job position. Our model achieves better F1-scores than previous approaches for each modality (verbal content, audio and video). Results from early and late multimodal fusion suggest that more sophisticated fusion schemes are needed to improve on the monomodal results. Finally, some examples of moments captured by the attention mechanisms suggest our model could potentially be used to help finding key moments in an asynchronous job interview.

\end{abstract}

\section{Introduction}

Among assessment methods, the job interview remains the most common way to evaluate candidates. The interview can be done via phone, live video, face to face, or more recently asynchronous video interview.
For the latter, candidates connect to a platform, and record themselves while answering a set of questions chosen by the recruiter. The platform then allows several recruiters to evaluate the candidate, to discuss among themselves and possibly to invite the candidate to a face-to-face interview. Recruiters choose to use these platforms because it gives them access to a larger pool of candidates, and it speeds up the application processing time. In addition, it allows candidates to do the interview whenever and wherever it suits them the most. However, given a large number of these asynchronous interviews it may quickly become unmanageable for recruiters. The highly structured characteristic of asynchronous video interviews (same questions, same amount of time per candidate) enhances their predictive validity, and reduces inter-recruiter variability  \cite{Schmidt2016TheFindings}. Moreover, recent advances in Social Signal Processing (SSP) \cite{Vinciarelli2014MoreComputing} have enabled automated candidate assessment \cite{Chen2017}, and companies have already started deploying solutions serving that purpose. However, previous studies used corpora of simulated interviews with limited sizes. The work proposed in this paper relies on a corpus that has been built in collaboration with a company and that consists of more than 7000 real job interviews for 475 open positions.
The size of this corpus enables the exploration of emerging models such as deep learning models, that are known to be difficult to deploy for Social Computing because of the difficulty to obtain large annotations of social behaviors. Based on those facts, we propose HireNet, a new hierarchical attention neural network for the purpose of automatically classifying candidates into two classes: \textit{hirable} and \textit{not hirable}. Our model aims to assist recruiters in the selection process. It does not aim to make any automatic decision about candidate selection. First, this model was built to mirror the sequential and hierarchical structure of an interview assessment: recruiters watch a sequence of questions and answers, which are themselves sequences of words or behavioral signals. 
Second, the HireNet model integrates the context of the open position (questions during the interview and job title) in order both to determine the relative importance between question-answer pairs and to highlight important behavioral cues with regard to a question. 
Third, HireNet  attention mechanisms enhance the interpretability of our model for each modality. In fact, they provide a way for recruiters to validate and trust the model through visualization, and possibly for candidates to locate their strengths or areas of improvement in an interview.\\
In this paper, we first present an overview of the related works for automatic video interview assessment. Then we go through the construction and the underlying hypotheses of HireNet, our neural model for asynchronous video interview assessment. After, we discuss the binary classification results of our model compared to various baselines, and show salient interview slices highlighted by the integrated attention mechanisms. Finally we conclude and discuss the future directions of our study.

\section{Related Work}

\subsection{Databases}

To the best of our knowledge, only one corpus of interviews with real open positions has been collected and is subject to automatic analysis \cite{Nguyen2014HireBehavior}. This corpus consists of face-to-face job interviews for a marketing short assignment whose candidates are mainly students. 
There are video corpora of face-to-face mock interviews that include two corpora built at the Massachusetts Institute of Technology \cite{Hoque2016Mach:Coach,Naim2018AutomatedPerformance}, and a corpus of  students in services related to hospitality \cite{Muralidhar2016TrainingHospitality}. Many corpora of simulated asynchronous video interviews have also been built: a corpus of employees \cite{Chen2016AutomatedParadigm}, a corpus of students from Bangalore University \cite{Rasipuram2017AutomaticInterviews} and a corpus collected through the use of crowdsourcing tools \cite{Chen2017}. Some researchers are also interested in online video resumes and have constituted a corpus of video CVs from YouTube \cite{Nguyen2016HirabilityResumes}. A first impressions challenge dataset was also supplemented by hirability annotation \cite{Escalante2017ChaLearnOverview}. Some corpora are annotated by experts or students in psychology \cite{Chen2016AutomatedParadigm,Chen2017,Nguyen2014HireBehavior,Rupasinghe2017ScalingAnalysis}. Other corpora have used crowdsourcing platforms or naive observers \cite{Rasipuram2017AutomaticInterviews} for annotation. Table\,\ref{previousCorpora} contains a summary of the corpora of job interviews used in previous works.

\begin{table*}
\centering
\begin{tabular}{|l|c|c|c|}
\hline
Works & Interview & Real open position & Number of candidates\\
\hline
\hline
\cite{Nguyen2014HireBehavior} & Face to Face & Marketing short assignment & 36 \\
\hline
\cite{Muralidhar2016TrainingHospitality} & Face to Face & None & 169 \\
\hline
\cite{Naim2018AutomatedPerformance} & Face to Face & None & 138 \\
\hline
\cite{Chen2016AutomatedParadigm} & Asynchronous Video & None & 36 \\
\hline
\cite{Rasipuram2017AutomaticInterviews} & Asynchronous Video & None & 106 \\
\hline
\cite{RaoS.B2017AutomaticStudy} & Asynchronous Video & None & 100 \\
\hline
\cite{Rupasinghe2017ScalingAnalysis} & Asynchronous Video & None & 36 \\
\hline
\cite{Chen2017} & Asynchronous Video & None & 260 \\
\hline
This Study & Asynchronous Video & Sales positions & 7095 \\
\hline
\end{tabular}%
\caption{Summary of job interview databases}
\label{previousCorpora}
\end{table*}

\subsection{Machine learning approaches for automatic analysis of video job interview}

\textbf{Features} Recent advances in SSP have offered toolboxes to extract features from audio \cite{Eyben2016TheComputing} and video streams \cite{Baltrusaitis2018OpenFaceToolkit}. As asynchronous job interviews are videos, features from each modality (verbal content, audio and video) have to be extracted frame by frame in order to build a classification model. 
Audio cues consist mainly of prosody features (fundamental frequency, intensity, mel-frequency cepstral coefficients, etc) and speaking activity (pauses, silences, short utterances, etc) \cite{Nguyen2015IMinute,RaoS.B2017AutomaticStudy}. Features derived from facial expressions (facial actions units, head rotation and position, gaze direction, etc) constitute the most extracted visual cues \cite{Chen2017}. Finally, advances in automatic speech recognition have enabled researchers to use the verbal content of candidates. In order to describe the verbal content, researchers have used lexical statistics (number of words, number of unique words, etc), dictionaries (Linguistic Inquiry Word Count) \cite{RaoS.B2017AutomaticStudy}, topic modeling \cite{Naim2018AutomatedPerformance}, bag of words or more recently document embedding \cite{Chen2016AutomatedParadigm}.

\textbf{Representation}
Once features are extracted frame by frame, the problem of temporality has to be addressed.
The most common approach is to simplify the temporal aspect by collapsing the time dimension using statistical functions (\textit{e.g.} mean, standard deviation, etc).
However, the lack of sequence modeling can lead to the loss of some important social signals such as emphasis by raising one's eyebrows followed by a smile \cite{Janssoone2016UsingStance}. Moreover co-occurrences of events are not captured by this representation. Thus, a distinction between a fake smile (activation of action unit 12) and a true smile (activation of action units 2, 4 and 12) is impossible \cite{Ekman1990TheII} without modeling co-occurrences.
To solve the problem of co-occurrences, the representation of visual words, audio words or visual audio words has been proposed \cite{Chen2017,Chen2016AutomatedParadigm,RaoS.B2017AutomaticStudy}.
The idea is to consider the snapshot of each frame as a word belonging to a specific dictionary. In order to obtain this codebook, an algorithm of unsupervised clustering is used to cluster common frames. Once we obtain the clusters, each class represents a "word" and we can easily map an ensemble of extracted frames to a document composed of these words. Then, the task is treated like a document classification. 
Additionally, the representation is not learned jointly with the classification models which can cause a loss of information.

\textbf{Modeling attempts and classification algorithms} As video job interviews have multiple levels, an architectural choice has to be made accordingly. Some studies tried to find the most salient moments during an answer to a question \cite{Nguyen2015IMinute}, the most important questions \cite{Naim2018AutomatedPerformance} or to use all available videos independently \cite{Chen2017} in order to predict the outcome of a job interview. Finally, when a sufficient representation is built, a classification or a regression model is trained. 
Regularized logistic regression (LASSO or Ridge), Random Forest and Support Vector Machines are the most widely used algorithms.

From a practical point of view, manually annotating thin slices of videos is time consuming. On the other side, considering each answer with the same label as the outcome of the interview is considerably less expensive, though some examples could be noisy. Indeed, a candidate with a negative outcome could have performed well on some questions.
Furthermore, all these models do not take into account the sequentiality of social signals or questions.

\subsection{Neural networks and attention mechanisms in Social Computing}

Neural networks have proven to be successful in numerous Social Computing tasks. Multiple architectures in the field of neural networks have outperformed hand crafted features for emotion detection in videos \cite{Zadeh2018}, facial landmarks detection \cite{Baltrusaitis2018OpenFaceToolkit}, document classification \cite{Yang2016} 
These results are explained by the capability of neural networks to automatically perform useful transformations on low level features. Moreover, some architectures such as Recurrent Neural Networks were especially tailored to represent sequences.
In addition, attention mechanisms have proven to be successful in highlighting salient information enhancing the performance and interpretability of neural networks.  
For example, in rapport detection, attention mechanisms allow to focus only on  important moments during dyadic conversations \cite{Yu2017TemporallyContent}. Finally, numerous models have been proposed to model the interactions between modalities in emotion detection tasks through attention mechanisms \cite{Zadeh2017TensorAnalysis,Zadeh2018}.

\section{Model}
\subsection{HireNet and underlying hypotheses}
We propose here a new model named HireNet, as in a neural network for hirability prediction. It is inspired by work carried out in neural networks for natural language processing and from the HierNet \cite{Yang2016}, in particular, which aims to model a hierarchy in a document. Following the idea that a document is composed of sentences and words, a job interview could be decomposed, as a sequence of answers to questions, and the answers, as a sequence of low level descriptors describing each answer.

\begin{figure}[!ht]
\centering
\includegraphics[width=1.\columnwidth]{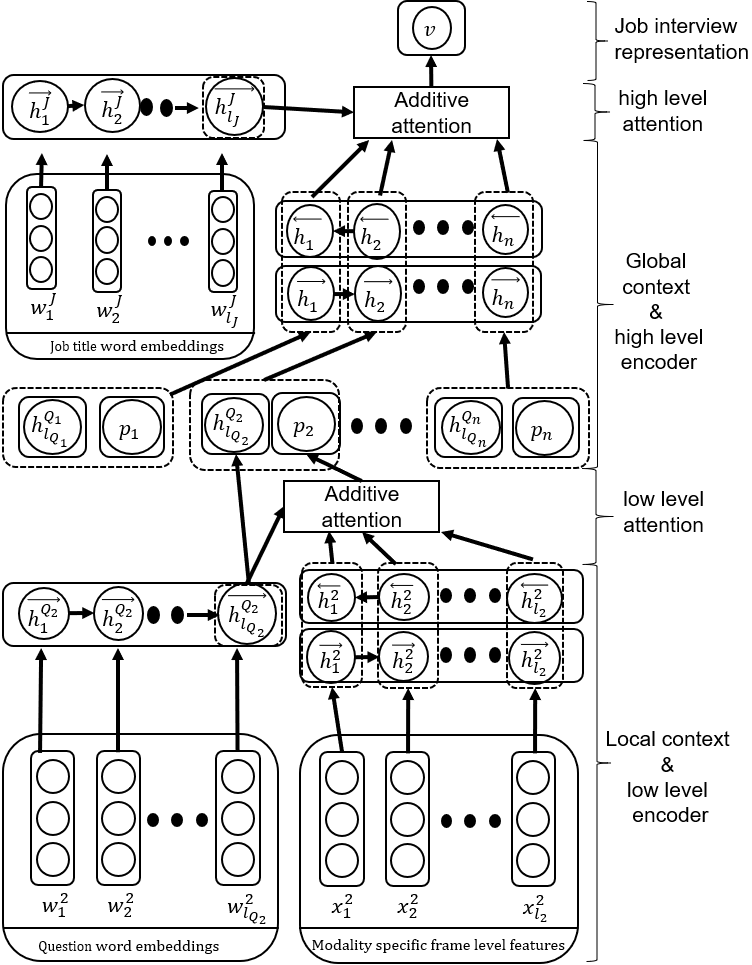}
\caption{HireNet}
\label{fig:HireNet}
\end{figure}

The model architecture (see Figure~\ref{fig:HireNet}) is built relying on four hypotheses. The first hypothesis (\textbf{H1}) is the importance of the information provided by the sequentiality of the multimodal cues occurring in the interview. We thus choose to use a sequential model such as a recurrent neural network. The second hypothesis (\textbf{H2}) concerns the importance of the hierarchical structure of an interview: the decision of to hire should be performed at the candidate level, the candidates answering several questions during the interview.  We thus choose to introduce different levels of hierarchy in HireNet namely the candidate level, the answer level and the word (or frame) level. The third hypothesis (\textbf{H3}) concerns the existence of salient information or social signals in a candidate's video interview: questions are not equally important and not all the parts of the answers have an equal influence on the recruiter's decision. We thus choose to introduce attention mechanisms in HireNet. The last hypothesis (\textbf{H4}) concerns the importance of contextual information such as questions and job titles. Therefore, HireNet includes vectors that encode this contextual information.

\subsection{Formalization}

We represent a video interview as an object composed of a job title $J$ and $n$ question-answer pairs \( 
\left \{ \left \{ Q_1,A_1 \right \},\left \{ Q_2,A_2 \right \}, \ldots,\left \{ Q_n,A_n \right \} \right \} \).
In our model, the job title $J$ is composed of a sequence of $l_J$ words \( \left \{ w^J_1, w^J_2, \ldots, w^J_{l_J} \right \} \) where $l_J$ denotes the length of the job title. In a same way, the $i$-th question $Q_i$ is a sequence of $l_{Q_i}$ words \( \left \{ w_1^{i}, w_2^{i}, \ldots, w^{i}_{l_{Q_i}} \right \} \) where $l_{Q_i}$ denotes the number of words in the question $i$. $A_i$ denotes the sequence of low level descriptors \( \left \{ x_1^{i}, x_2^{i}, \ldots, x^{i}_{l_{A_i}} \right \} \) describing the $i$-th answer. In our study these low level descriptors could be embedded words, features extracted from  an audio frame, or features extracted from a video frame. $l_{A_i}$ denotes the length of the sequence of low level 
descriptors of the $i$-th answer. 
\subsubsection{Gated Recurrent Unit Encoder}

We decided to use a Gated Recurrent Unit (GRU)  \cite{cho2014learning} to encode information from the job title, the questions and the answers. A GRU is able to encode sequences. It uses two mechanisms to solve the vanishing gradient problem, namely the reset gate, controlling how much past information is needed; and the update gate, determining how much past information has to be kept and the amount of new information to add. 
For formalization, we will denote by $h_{t}$ the hidden state of GRU at timestep $t$ of the encoded sequence.

\subsubsection{Low level encoder}

This part of the model aims to encode the sequences of low level descriptors. As mentioned before, the sequences can represent a text, an audio stream or a video stream. A bidirectional GRU is used to obtain representations from both directions for each element of the sequence $X$. It contains the forward \overrightarrow{GRU} which reads the sequence from left to right and backward \overleftarrow{GRU} which reads the sequence from right to left: 
\begin{equation}
 \overrightarrow{h_{t}^i} = \overrightarrow{GRU}(x_t^{i}) , t \in [1,l_{A_i} ]
\end{equation}

\begin{equation}
 \overleftarrow{h_{t}^i} = \overleftarrow{GRU}(x_t^{i}) , t \in [l_{A_i}, 1]
\end{equation}

In the same way, an encoding for a given low level descriptor $x_t^{i}$ is obtained by concatenating forward hidden states and backward hidden states:
\begin{equation}
 h_{t}^i = [ \overrightarrow{h_{t}^i}, \overleftarrow{h_{t}^i}]
\end{equation}

Encoding sequences in a bidirectional fashion ensures the same amount of previous information for each element of $(A_i)_{1\leq i\leq n}$. Using a simple forward encoder could lead to biased attention vectors focusing only on the latest elements of the answers.

\subsubsection{Local context encoder}

In this study, the local context information corresponds to the questions $(Q_i)_{1\leq i\leq n}$. In order to encode these sentences, we use a simple forward GRU.
\begin{equation}
\overrightarrow{h_{t}^{Q_i}} = \overrightarrow{GRU}(w_t^{i}) , t \in [1,l_{Q_i} ]
\end{equation}
And the final representation of a question is the hidden state of the last word in the question $Q_i$ (\textit{i.e.} $h_{l_{Q_i}}^{Q_i}$).

\subsubsection{Low level attention}

In order to obtain a better representation of of the candidate's answer, we aim to detect elements in the sequence which were salient for the classification task. Moreover, we hypothesize that the local context is highly important. Different behavioral signals can occur depending on the question type and it can also influence the way recruiters assess their candidates \cite{Roulin2015HonestEvaluations}. An additive attention mechanism is proposed in order to extract the importance of each moment in the sequence representing the answer.

\begin{equation}
 u_{t}^{i} = tanh(W_A h_{t}^i + W_Q h_{l_{Q_i}}^{Q_i}+b_Q)
\end{equation}
\begin{equation}
 \alpha^{i}_t = \frac{exp(u_p^\top {u_{t}^{i}} )}{\sum\nolimits_{t'} exp(u_p^\top {u_{t'}^{i}})}
\end{equation}
\begin{equation}
 a_i = \sum\nolimits_{t} \alpha^{i}_t h_{t}^i 
\end{equation}

where $W_A$ and $W_Q$ are weight matrices, $u_p$ and $b$ are weight vectors and $u_p^\top$ denotes the transpose of  $u_p$.

\subsubsection{High level encoder}

In order to have the maximum amount of information, we concatenate at the second level, the representation of the local context and the answer representation. 
Moreover, we think that given the way video interviews work, the more questions a candidate answers during the interview, the more he adapts and gets comfortable. In the light of this, we decided to encode question-answer pairs as a sequence. Given \( \left \{     [ \ h_{l_{Q_1}}^{Q_1}, a_1 ] \ ,[ \ h_{l_{Q_2}}^{Q_2}, a_2 ] \ , \ldots, [ \ h_{l_{Q_n}}^{Q_n}, a_n ] \   \right \} \), we can use the same representation scheme as that of the low level encoder:

\begin{equation}
 \overrightarrow{h_i} = \overrightarrow{GRU}([ h_{l_{Q_i}}^{Q_i}, a_i ]) , i \in [1,n ]
\end{equation}

\begin{equation}
 \overleftarrow{h_i} = \overleftarrow{GRU}([ h_{l_{Q_i}}^{Q_i}, a_i ]) , i \in [n,1 ]
\end{equation}

We will also concatenate forward hidden states and backward hidden states:
\begin{equation}
 h_i = [ \overrightarrow{h_i}, \overleftarrow{h_i}]
\end{equation}

\subsubsection{Global context encoder}

We encode the job title the same way we encode the questions :

\begin{equation}
\overrightarrow{h_{t}^{J}} = \overrightarrow{GRU}(w_t^{J}) , t \in [1,l_J]
\end{equation}
As done for the representation of the question, the final representation of the job title is the hidden state of the last word of $J$ (\textit{i.e.} $h_{l_J}^{J}$).

\subsubsection{High level attention}

The importance of a question depends on the context of the interview, and specifically, on the type of job the candidate is applying for. For instance, a junior sales position interview could accord more importance to the social skills, while an interview for a senior position could be more challenging on the technical side.\\
Like low level attention, high level attention is composed of an additive attention mechanism:

\begin{equation}
 u_{i} = tanh(W_P h_i + W_J h_{l_J}^{J}+b_J)
\end{equation}
\begin{equation}
 \alpha_i = \frac{exp(u_J^\top {u_{i}} )}{\sum\nolimits_{i'} exp(u_J^\top {u_{i'}})}
\end{equation}
\begin{equation}
 v = \sum\nolimits_{i} \alpha_i h_i 
\end{equation}
where $W_P$, $W_J$ are weight matrices, $u_J$ and $b_J$ are weight vectors and $u_J^\top$ denotes the transpose of  $u_J$.
Finally $v$ summarizes all the information of the job interview.

\subsubsection{Candidate classification}

Once $v$ is obtained, we use it as representation in order to classify candidates:
\begin{equation}
 \widetilde{y} = \sigma(W_v v + b_v)
\end{equation}
where $W_v$ is a weight matrix and $b_v$ a weight vector.
As the problem we are facing is that of a binary classification, we chose to minimize the binary cross-entropy computed between $\widetilde{y}$ and true labels of candidates $y$.

\section{Experiments}

\subsection{Dataset}
We have decided to focus on only one specific type of job: sales positions.
After filtering based on specific job titles from the ROME Database\footnote{https://www.data.gouv.fr/en/datasets/repertoire-operationnel-des-metiers-et-des-emplois-rome/}, a list of positions was selected and verified by the authors and an expert from the Human Resources (HR). Finally, in a collaboration with an HR industry actor, we have obtained a dataset of French video interviews  comprising more than 475 positions and 7938 candidates. As they watch candidates' videos, recruiters can like, dislike, shortlist candidates, evaluate them on predefined criteria, or write comments. To simplify the task, we set up a binary classification: candidates who have been liked or shortlisted are considered part of the \textit{hirable} class and others part of the \textit{not hirable} class. If multiple annotators have annotated the same candidates, we proceed with a majority vote. In case of a draw, the candidate is considered \textit{hirable}. It is important to note that the videos are quite different from what could be produced in a laboratory setup. Videos can be recorded from a webcam, a smartphone or a tablet., meaning noisy environments and low quality equipment are par for the course. Due to these real conditions, feature extraction may fail for a single modality during a candidate's entire answer. One example is the detection of action units when the image has lighting problems. We decided to use all samples available in each modality separately.
Some statistics about the dataset are available in Table \ref{table:datasets}. Although the candidates agreed to the use of their interviews,  the dataset will not be released to public outside of the scope of this study due to the videos being personal data subject to high privacy constraints.

\begin{table}
\begin{tabular}{|l|c|c|c|}
\hline
Modality                                                             & Text         & Audio   & Video   \\
\hline
\hline
Train set     & 6350         & 6034    & 5706    \\
\hline
Validation set  & 794          & 754     & 687     \\
\hline
Test set      & 794          & 755     & 702     \\
\hline
\hline
\begin{tabular}[c]{@{}l@{}}Questions per\\ interview (mean)\end{tabular} & 5.05         & 5.10    & 5.01    \\
\hline
Total length                                                        & 3.82\,M\,words & 557.7\,h & 508.8\,h \\
\hline
\begin{tabular}[c]{@{}l@{}}Length per \\ question (mean)\end{tabular} & 95.2\,words   & 52.19\,s & 51.54\,s \\
\hline
\begin{tabular}[c]{@{}l@{}}\textit{Hirable} label\\ proportion\end{tabular}  & 45.0\,\%         & 45.5\,\%    & 45.4\,\%   \\
\hline
\end{tabular}
\caption{Descriptive table of the dataset: number of candidates in each set and overall statistics of the dataset.}
\label{table:datasets}
\end{table}

\subsection{Experimental settings}

\begin{table*}[t!]
\centering
\begin{tabular}{|l|c|c|c||c|c|c||c|c|c|}
\hline
 & \multicolumn{3}{c||}{Text} & \multicolumn{3}{c||}{Audio} & \multicolumn{3}{c|}{Video}\\ \hline
 Model &  $Precision$ & $Recall$& $\textbf{F1}$ & $Precision$ & $Recall$& $\textbf{F1}$&  $Precision$ & $Recall$& $\textbf{F1}$\\ \hline
Non-sequential& 0.553 & 0.285 & 0.376 &0.590 & 0.463& 0.519& 0.507 & 0.519& 0.507\\ \hline
Bo*W&0.656 & 0.403& 0.499& 0.532 & 0.402& 0.532&0.488 & 0.447& 0.467\\ \hline  \hline
Bidirectional GRU&0.624& 0.510& 0.561& 0.539& 0.596& 0.566& 0.559& 0.500& 0.528\\ \hline
HN\_AVG&0.502& 0.800& 0.617&0.538& 0.672& 0.598&0.507& 0.550& 0.528\\ \hline
 HN\_SATT& 0.512& 0.803& 0.625& 0.527& 0.736& 0.614& 0.490& 0.559& 0.522\\ \hline
 HireNet &0.539& 0.797& \textbf{0.643}&0.576& 0.724& \textbf{0.642}&0.562& 0.655& \textbf{0.605} \\ \hline  
 \end{tabular}
\caption{Results for Monomodal models}
\label{ResultsTable}
\end{table*}

The chosen evaluation metrics are precision, recall and F1-score of \textit{hirable} class. They are well suited for binary classification and used in previous studies \cite{Chen2017}. We split the dataset into a training set, a validation set for hyper-parameter selection based on the F1-score, and a test set for the final evaluation of each model. Each set constitutes respectively 80\%, 10\% and 10\% of the full dataset.

\subsection{Extraction of social multimodal features}
For each modality, we selected low-level descriptors to be used as per-frame features, and sequence-level features to be used as the non-sequential representation of a candidate's whole answer for our non-sequential baselines.

\textbf{Word2vec}: Pretrained word embeddings are used for the BoTW (Bag of Text Words, presented later in this section), and the neural networks. We used word embeddings of dimension 200 from \cite{fauconnier_2015} pretrained on a French corpus of Wikipedia.

\textbf{eGeMAPS}: Our frame-level audio features are extracted using OpenSmile \cite{Eyben2013b}. The configuration we use is the same one used to obtain the eGeMAPS \cite{Eyben2016TheComputing} features. GeMAPS is a famous minimalistic set of features selected for their saliency in Social Computing, and eGeMAPS is its extended version. We extract the per-frame features prior to the aggregations performed to obtain the eGeMAPS representation.

\textbf{OpenFace}:
We extract frame-level visual features with OpenFace \cite{Baltrusaitis2018OpenFaceToolkit}, a state-of-the-art visual behavioral analysis software that yields various per-frame meaningful metrics. We chose to extract the position and rotation of the head, the intensity and presence of actions units, and the gaze direction. As different videos have different frame-rates, we decided to smooth values with a time-window of 0.5\,s and an overlap of 0.25\,s. The duration of 0.5\,s is frequently used in the literature of Social Computing \cite{Varni2018ComputationalInteractions} and has been validated in our corpus as a suitable time-window size by annotating segments of social signals in a set of videos.
\subsection{Baselines}
First, we compare our model with several vote-based methods: \textit{i)} \textbf{Random vote baseline} (One thousand random draws respecting the train dataset label balance were made. The F1-score is then averaged over those one thousand samples); \textit{ii)} \textbf{Majority Vote} (This baseline is simply the position-wise majority label. Since our model could just be learning the origin open position for each candidate and its corresponding majority vote, we decided to include this baseline to show that our model reaches beyond those cues).\\
Second, we compare our model with non-sequential baselines: 
\textit{i)-a} \textbf{Non-sequential text} (we train a Doc2vec \cite{DBLP:conf/icml/LeM14} representation on our corpus, and we use it as a representation of our textual inputs);
\textit{i)-b} \textbf{Non-sequential audio} (we take the eGeMAPS audio representation as described in \cite{Eyben2016TheComputing}. That representation is obtained by passing the above descriptors into classical statistical functions and hand-crafted \emph{ad hoc} measures applied over the whole answer. The reason we chose GeMAPS features is also that they were designed to ease comparability between different works in the field of Social Computing);
\textit{i)-c} \textbf{Non-sequential video} (our low-level video descriptors include binary descriptors and continuous descriptors.  The mean, standard deviation, minimum, maximum, sum of positive gradients and sum of negative gradients have been successfully used for a behavioral classification on media content in \cite{Ryoo2015PooledVideos}. We followed that representation scheme for our continuous descriptors. As for our discrete features, we chose to extract the mean, the number of active segments, and the active segment duration mean and standard deviation)
\textit{ii)} \textbf{Bag of * Words} (We also chose to compare our model to  \cite{Chen2017}'s Bag of Audio and Video Words: we run a K-means algorithm on all the low-level frames in our dataset. Then we take our samples as documents, and our frames' predicted classes as words, and use a "Term Frequency-inverse Document Frequency" (TF-iDF) representation to model each sample).
\\
For each modality, we use the non-sequential representations mentioned above in a monomodal fashion as inputs to three classic learning algorithms (namely SVM, Ridge regression and Random Forest) with respective hyperparameter searches. Best of the three algorithms is selected. As these models do not have a hierarchical structure, we will train them to yield answer-wise labels (as opposed to the candidate-wise labeling performed by our hierarchical model). At test time we average the output value of the algorithm for each candidate on the questions he answered.\\
Third, the proposed sequential baselines aim at checking the four hypotheses described above: \textit{i)} comparing the \textbf{Bidirectional-GRU model} with previously described non sequential approaches aims to validate \textbf{H1} on the contribution of sequentiality in an answer-wise representation; \textit{ii)} the Hierarchical Averaged Network (HN\_AVG) baseline adds the hierarchy in the model in order to verify \textbf{H2} and \textbf{H3} (we replace the attention mechanism by an averaging operator over all of the non-zero bidirectional GRU outputs); \textit{iii)} the Hierarchical Self Attention Network (HN\_SATT) is a self-attention version of HireNet which aims to see the actual effect of the added context information (\textbf{H4}).
\subsection{Multimodal models}
Given the text, audio, and video trained versions of our HireNet, we report two basic models performing multimodal inference, namely an early fusion approach and a late fusion approach. In the early fusion, we concatenate the last layer $v$ of each modality as a representation 
,and proceed with the same test procedure as our non-sequential baselines. For our late fusion approach, the decision for a candidate is carried out using the average decision score $\widetilde{y}$ between the three modalities.

\section{Results and analyses}

\begin{table}[htb!]
\centering
\begin{tabular}{|l|c|c|c|}
\hline
 Model &  $Precision$ & $Recall$& $\textbf{F1}$ \\ \hline
Random vote& 0.459 & 0.452 & 0.456\\ \hline
 Majority vote&0.567 & 0.576 & 0.571\\ \hline \hline
 Early Fusion&0.587& 0.705 & 0.640\\ \hline
  Late Fusion&0.567 & 0.748 & \textbf{0.645}\\ \hline
 \end{tabular}
\caption{Results for Multimodal models and vote-based baselines}
\label{ResultsTable2}
\end{table}

First of all, Tables \ref{ResultsTable} and \ref{ResultsTable2} show that most of our neural models fairly surpass the vote-based baselines. \\ 
In Table \ref{ResultsTable}, the F1-score has increased, going from the non-sequential baselines, to the Bidirectional-GRU baselines for all the modalities, which supports \textbf{H1}. We can also see that HN\_AVG is superior to  the Bidirectional-GRU baselines for audio and text validating \textbf{H2} for those two modalities.
This suggests that sequentiality and hierarchy are adequate inductive biases for a job interview assessment machine learning algorithm.
As for \textbf{H3}, HN\_SATT did show better results than HN\_AVG, for text and audio. 
In the end, our HireNet model surpasses HN\_AVG and HN\_SATT for each modality. Consequently, a fair amount of useful information is present in the contextual frame of an interview, and this information can be leveraged through our model, as it is stated in \textbf{H4}.
Audio and text monomodal models display better performance than video models. The same results were obtained in \cite{Chen2017}.\\
Our attempts at fusing the multimodal information synthesized in the last layer of each HireNet model only slightly improved on the single modality models.  
\subsection{Attention visualization}

\textbf{Text} In order to visualize the different words on which attention values  were high, we computed new values of interest as it has been done in \cite{Yu2017TemporallyContent}. As the sentence length changes between answers, we multiply every word's attention value ($\alpha_t^i$) by the number of words in the answer, resulting in the relative attention of the word with respect to the sentence. 
In a same way, we multiply each question attention by the number of questions, resulting in the relative attention of the question with respect to the job interview. Then, in a similar way as \cite{Yang2016}, we compute $\sqrt{p_q}p_w$ where $p_w$ and $p_q$ are respectively the values of interest for word $w$ and question $q$.  
The list of the 20 most important words contains numerous names of banks and insurances companies (Natixis, Aviva, CNP, etc) and job knowledge vocabulary (mortgage, brokerage, tax exemption, etc), which means that their occurrence in candidates answers takes an important role in hirability prediction. 

\begin{figure}[htb!]
\centering
  \includegraphics[width=\columnwidth]{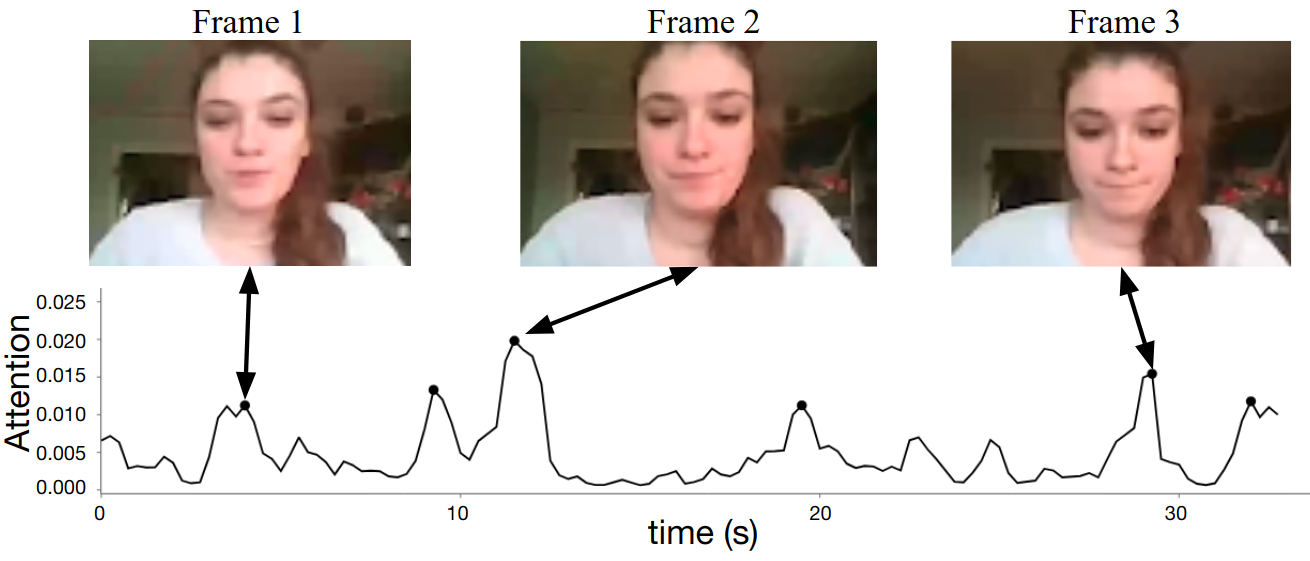}
  \caption{Example of salient moments detected with peaks of attention on the video modality}
  \label{video}
\end{figure}
\textbf{Video} In order to visualize which moments were highlighted by attention mechanisms in a video, we display an example of the attention values for an answer in Figure \ref{video}. 
In this figure, the higher the attention value, the more the corresponding frames are considered task-relevant by the attention mechanism. As we can see, some peaks are present. Three thin slices with high attention values are presented.
Some social signals that are important in a job interview are identified. We hypothesize that the smile detected in Frame 1 could be part of a tactic to please the interviewer known as deceptive ingratiation \cite{Schneider2015CuesInterview}. In addition, Frames 2 and 3 are representative of stress signals from the candidate. In fact, lip suck was suggested to be linked to anxiety in \cite{Feiler2016BehavioralAnxiety}.

\textbf{Audio} The same visualization procedure used for video has been investigated for audio. As audio signal is harder to visualize, we decided to describe the general pattern of audio attention weights. In most cases, when the prosody is homogeneous through the answer, attention weights are distributed uniformly and show no peaks, as opposed to what was observed for video. However, salient moments may appear, especially when candidates produce successive disfluencies. Thus, we have identified peaks where false starts, filler words, repeating or restarting sentences occur.

\begin{figure}
\centering
  \includegraphics[width=0.77\columnwidth]{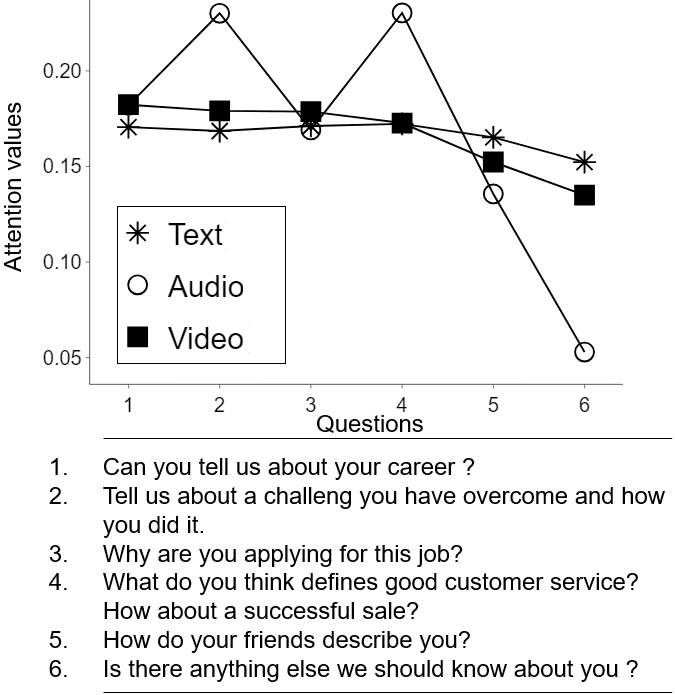}
  \caption{Questions describing the randomly picked open position and their respective attention values }
  \label{questionsatt}

\end{figure}

\textbf{Questions} We aim to explore the attention given to the different questions during the same interview. For this purpose, we randomly picked one open position from the test dataset comprising 40 candidates. Questions describing the interview and the corresponding averaged attention weights are displayed in the Figure \ref{questionsatt}. First, it seems attention weight variability between questions is higher for the audio modality than for text and video modalities. 
Second, the decrease in attention for Questions 5 and 6 could be explained by the fact that those questions are designed to assess "soft skills".
Third, peaks of attention weight for the audio modality on Questions 2 and 4 could be induced by the fact that these questions are job-centric. Indeed, it could be possible that disfluencies tend to appear more in job-centric questions or that prosody is more important in first impressions of competence.

\section{Conclusion and future directions}

The HR industry actors nowadays do offer tools to automatically assess candidates undergoing asynchronous video interviews. However, no studies have been published regarding these tools and their predictive validity. 
The contribution of this work is twofold. First, we evaluate the validity of previous approaches in real conditions (\textit{e.g.} in-the-wild settings, true applications, real evaluations, etc). Second, we used deep learning methods in order to faithfully model the structure of asynchronous video interviews.
In that sense, we proposed a new version of Hierarchical Attention Networks that is aware of the interviews contextual elements (questions and job title) called HireNet, which has showed better performance than previous approaches. First basic experiments on multimodal fusion have also been performed (early and late fusion). In future work, the obtained multimodal performance could be improved by leveraging more sophisticated multimodal fusion schemes.  HireNet was evaluated on a corpus containing interviews for various jobs -- 475 different positions -- in the domain of sales positions. Theoretical findings from industrial organizational psychology suggest that some dimensions are common across different positions\cite{Huffcutt2001IdentificationInterviews.}. However we would like to extend the corpus to other domains than sales in order to i) validate the relevance of our model for other types of positions, ii) determine which competencies are common or not across jobs. In that sense, the use of multi-domain models\cite{Liu2017AdversarialClassification} could be of great help. Our model currently considers two labels (“hirable” and “not hirable”). Extending our annotations to more fine-grained information (communication skills, social effectiveness, etc) could  provide useful insights about the profile of a candidate and its potential fit with the position in question.
Through the use of attention mechanisms, we aimed to highlight salient moments and questions for each modality, which contributes to the transparency and the interpretability of HireNet. Such transparency is very important for Human Resources practitioners to trust an automatic evaluation. Further investigations could be conducted on the proposed attention mechanisms: i) to confirm the saliency of the selected moments using the discipline of Industrial and Organizational psychology; ii) to know the influence of the slices deemed important. This way, a tool to help candidates train for interviews could be developed.

 Last but not least, ethics and fairness are important considerations, that deserve to be studied. In that sense, detection of individual and global bias should be prioritized in order to give useful feedbacks to practitioners. Furthermore we are considering using adversarial learning as in \cite{Zhang2018} in order to ensure fairness during the training process.

\section{Acknowledgments}

This work was supported by the company EASYRECRUE, from whom the job interview videos were collected. We would like to thank Jeremy Langlais for his support and his help. We would also like to thank Valentin Barriere for his valuable input and the name given to the model and Marc Jeanmougin and Nicolas Bouche for their help with the computing environment. Finally, we thank Erin Douglas for proofreading the article.

\bibliographystyle{aaai} \bibliography{Mendeley}

\end{document}